\documentclass[11pt,a4paper]{article}

\usepackage[utf8]{inputenc}
\usepackage[T1]{fontenc}
\usepackage[margin=1in]{geometry}
\usepackage{amsmath,amssymb}
\usepackage{graphicx} 
\usepackage{booktabs}
\usepackage{array} 
\usepackage{hyperref}
\usepackage{natbib} 
\usepackage{xcolor}
\usepackage{caption}
\usepackage{float}

\hypersetup{
	colorlinks=true,
	linkcolor=blue,
	filecolor=magenta,      
	urlcolor=cyan,
	citecolor=blue,
}

\title{\textbf{Machine Learning-Based Detection of Mobile Money Scam Messages in Ghana: \\ A Comparative Evaluation of Na\"ive Bayes, Support Vector Machine, and Random Forest Classifiers}}

\author{
	Ibrahim Salifu\(^{1}\) \and Muhaisin Tiyumba Nantogmah\(^{2}\) \\
	\vspace{0.5em}
	\small \(^{1}\)Department of Computer Science, University of Technology and Applied Sciences, Navrongo, Ghana \\
	\small \texttt{sibrahim26.stu@utas.edu.gh} \\
	\vspace{0.5em}
	\small \(^{2}\)Department of Computer Science, University for Development Studies, Tamale, Ghana \\
	\small \texttt{muhaisintiyumbana1848@uds.edu.gh}
}
\date{}

\begin{document}
	
	\maketitle
	
	\begin{abstract}
		\noindent Mobile money (MoMo) has emerged as the preferred payment channel for retail transactions in Ghana, accounting for a record GH\c{5}518.4 billion worth of transactions in December 2025 alone, and with over 26.7 million active accounts across the country. In tandem with this explosive growth, the same growth has occurred also in social-engineering fraud over Short Message Service (SMS) and Unstructured Supplementary Service Data (USSD) channels, where the fraudsters pretend to be network operators, agents, relatives or regulators and elicit victims to reveal their personal identification numbers (PINs) and/or identity and/or transfer funds. In this study, we propose a supervised machine learning pipeline to discriminate between scam and legitimate Ghanaian mobile money SMS messages in a labelled corpus of 220 SMS messages from 18 different sub-categories of scam and legitimate messages. Pre-processing domain-specific normalisation for currency amounts, phone numbers and USSD codes was done, and vectorised using term frequency-inverse document frequency (TF-IDF) weighted unigrams and bigrams. Stratified five-fold cross validation with a held out test set of 20\% of the data was used to train and evaluate three classifiers: Multinomial Na\"ive Bayes, linear Support Vector Machine (SVM), and Random Forest. These classifiers were evaluated using accuracy, precision, F1-score, Matthews Correlation Coefficient (MCC), and Area Under the Receiver Operating Characteristic Curve (AUC-ROC). The held-out test performance (accuracy, F1 and AUC) showed perfect performance (accuracy = 1.00, F1 = 1.00 and AUC = 1.00) for Naive Bayes and linear SVM, and averaged \(96.6\%\) (\(\pm 3.3\)), \(94.9\%\) (\(\pm 4.6\)) and \(92.0\%\) (\(\pm 4.2\)), respectively for cross-validated accuracy over folds for Naive Bayes, SVM and Random Forest. The most discriminative n-grams were found in the imperative urgency words used in scam messages (e.g., ``send'', ``claim'', ``please send'') and the PIN-solicitation words used in the scam messages (e.g., ``PIN'', ``pincode''). In legitimate messages, the words, such as ``balance'', ``successful'', and ``do not share'', concentrated on confirmatory and cautionary language. We put these results in the context of the broader body of work on fraud detection using SMS in Sub-Saharan Africa, and consider the ceiling effect of small and template-based corpora, and suggest a pathway toward larger, naturalistic, multilingual corpora. The results show that lightweight and interpretable classical classifiers show high discriminative accuracy on structured mobile money scam text and are well suited to be deployed in carrier level or handset level filtering systems in low-resource settings on device.
		
		\vspace{1em}
		\noindent \textbf{Keywords:} mobile money fraud, SMS scam detection, smishing, machine learning, Naive Bayes, support vector machine, random forest, TF-IDF, Ghana, and fintech security.
	\end{abstract}
	
	\section{Introduction}
	It has been seen that mobile money has transformed the financial inclusion of the unbanked and underbanked people in Sub-Saharan Africa, allowing them to hold some value, send money, and make payments using a mobile phone without having a formal bank account \citep[as cited by][]{lokanan2023predicting,aron2018mobile}. This is because the total annual transaction value of the mobile money sector in Ghana reached GH\c{5}1 trillion in 2022, jumped to a provisional GH\c{5}4.5 trillion in 2025, with GH\c{5}518.4 billion worth of transactions expected to be recorded in December 2025, with 982 million individual transactions and 26.7 million active mobile money accounts expected to be recorded \citep{graphic2026,nbc2026}. In June 2026, the total number of registered accounts was still over 84 million, and the total value of transactions was GH\c{5}492.9 billion per month \citep{gna2026}. This adoption is changing the dynamics of financial inclusion and it is creating a huge and lucrative opportunity for fraudsters to attack.
	
	Unlike the Western world, where the mobile money market is heavily targeted, the majority of attacks against mobile money in Ghana are through SMS, voice call and informal agents interaction \citep{akomea2019control}. The majority of scam scripts are about a ``cash-out,'' requiring the money to be sent ``upfront'' in order to be ``activated,'' or a bank transfer having been credited to the wrong person and must be ``reversed,'' or there's a prize or lottery, or there's an emergency and someone needs the funds from relatives \citep{bft2026}. The use of generative artificial intelligence (AI) to generate many scam messages in local languages like Twi is becoming more common as fraud operations are increasingly professionalising in Ghana, and the lack of awareness of mobile users' cyber-literacy may be a growing problem for them \citep{mobileidworld2025}.
	
	The present Ghanaian measures to mitigate the use of SMS for mobile money fraud are mostly reactive and customer-driven, such as public awareness creation via SMS and message complaints received by the customers after the fraudulent transaction by the operators. There is significantly less literature available on how supervised machine learning could be directly applied to the text of SMS spam messages originating from mobile money services in Ghana for the automatic and real-time filtering of SMS spam. The few studies found in the literature have focused on generic SMS spam detection over English-language corpora, such as the well-known UCI SMS Spam Collection \citep{almeida2011contributions} and large scale SMS spam reporting databases from the United Kingdom \citep{agarwal2025overview} and Australia \citep{salman2022empirical}. Previous studies focusing on the African context have mostly focused on anomaly detection at the transaction level, either based on structured numerical features \citep{lokanan2023predicting}, or on other national context and language, such as the Chichewa SMS fraud \citep{chavula2025using}, or on the e-wallet deposit-notification scam in Namibia \citep{enkono2020application}.
	
	In this paper, we bridge this gap by developing and thoroughly evaluating a text-classification pipeline for Ghanaian mobile money SMS messages using a manually curated and labelled dataset with all transaction alerts, one-time-password (OTP) notices, loan scams, fake prize and refund scams, agent-impersonation scams, family-emergency scams, investment frauds and eight other subtitles. This study has made four contributions:
	
	\begin{enumerate}
		\item Canonicalisation of currency amounts (Ghanaian currency figures: GHC/GHS), local phone number formats and USSD dial-strings (e.g., *170\#) before vectorising Ghanaian mobile money SMS, enhancing the signal-to-noise ratio of the resulting feature space.
		\item A systematic comparison of three classical classifiers (Multinomial Na\"ive Bayes, linear Support Vector Machine (SVM), and Random Forest) based on TF-IDF unigram/bigram features, with a battery of metrics (accuracy, precision, recall, F1, MCC and AUC-ROC) for both a stratified k-fold cross validation and a held-out test partition.
		\item An interpretability analysis which identifies lexical and structural markers that are most correlated with scam intent and relates these markers to well-documented social-engineering persuasion tactics (urgency, authority impersonation, and reciprocity).
		\item A clear and rigorous presentation of the ceiling-effect issue of small, template-type datasets, followed by a sensible scheme for scaling this work up to a naturalistic, multilingual and adversarial robust deployment-ready system.
	\end{enumerate}
	
	The rest of the paper is organised as follows. The related work carried out in SMS spam detection and mobile money fraud detection is reviewed in Section \ref{sec:related}. Section \ref{sec:methods} provides details about the dataset, the preprocessing pipeline, feature engineering, and the classification models. The experimental results are provided in Section \ref{sec:results} in the form of confusion matrixes, ROC curves and feature-Importance plots. In Section \ref{sec:discussion} the implications, novelty and limitations of the findings are discussed, and Section \ref{sec:conclusion} ends the research with directions for future research.
	
	\section{Related Work}
	\label{sec:related}
	
	\subsection{Mobile Money Fraud in Ghana and Sub-Saharan Africa}
	\citet{akomea2019control} identified weak know your customer (KYC), agent level collusion, and low consumer fraud-literacy as some of the first exploratory studies to uncover structural weaknesses in the Ghanaian mobile money service. More recent research than above has looked at the problem of detecting mobile money fraud as a data-mining problem on top of transaction records, with a proposed hybrid machine-learning framework for fraud alerting in Ghanaian mobile money services relying on the corpus of historical frauds, and other research that uses structured mobile money transaction data including Support Vector Machine, gradient-boosted decision tree and Naive Bayes algorithms, where gradient-boosted decision trees achieve particularly good fraud detection performance for structuring ``smurfing'' behaviour \citep{appiah2022fraud}. \citet{lokanan2023predicting} built upon this per-transaction model with its larger class of supervised fraud machine learning, within the financial-crime literature on mobile money and financial inclusion \citep{aron2018mobile} and the mobile money cost-sensitive fraud detection literature originally developed for credit-card fraud \citep{almahithawi2020example}.
	
	These approaches, which are complementary to the text-classification approach used in the present study, target numerical data pertaining to the transactions (amounts, time stamps, network graphs) but do not focus on the natural-language content of the SMS messages that is used to carry out the initial social-engineering deception. Recent headlines in Ghana's financial press \citep{bft2026} and reports of scammers using AI-generated scripts in the local language to time attacks \citep{mobileidworld2025} all point to the fact that most scam attacks these days are performed via the deception layer.
	
	\subsection{SMS Spam and Smishing Detection with Classical Machine Learning}
	The UCI SMS Spam Collection gathered by \citet{almeida2011contributions} has been used as a benchmark for SMS-text classification and has served as a good reference point for spam txt/ham txt classification since then, as it introduced the use of TF-IDF feature extraction with Naive Bayes and Support Vector Machine classifiers as a strong baseline. However, a subsequent comparison of TF-IDF with the bag of words approach over various families of classifiers has consistently shown that the TF-IDF approach improves the performances of all the classifiers when compared to the bag of words approach by itself. A recent large-scale comparison of six classifiers \citep{arxiv2025leveraging} for a standard SMS spam benchmark showed that Na\"ive Bayes with TF-IDF achieved \(96.2\%\) accuracy and linear SVM with TF-IDF achieved \(94.5\%\) accuracy, both outperforming K-Nearest Neighbours, Linear Discriminant Analysis, and Decision Trees. \citet{salman2022empirical} continued this trend by adding a fresh dataset of 153,551 SMSs to the scam set and also examined the robustness of the models to adversarial text disruption, demonstrating that while shallow classical classifiers performed well on the un-perturbed problem they were significantly more vulnerable to adversarial evasion -- which is pertinent for any deployed scam-filtering system.
	
	It is also possible to gain insights into the structure of the scams that classifiers need to learn to detect, by conducting large-scale measurement studies of real-world smishing. Analysing 1.35 million SMS messages reported by users to a major UK mobile network operator's ``7726'' SMS reporting service, \citet{agarwal2025overview} identified \(40.27\%\) of the reports as scams (excluding unsolicited but not malicious spam) and classified them into twelve different types, all of which broadly align with the existing taxonomy of ``agent-impersonation'', ``family-emergency'', and ``fake-reversal'' scams identified in the present Ghanaian data. In the mobile-money specific literature, \citet{enkono2020application} tried to detect fraudulent e-wallet deposit-notification SMS messages in Namibia using Na\"ive Bayes and SVM classifiers, which were essentially the same categories analysed here, but also with a dataset of machine-translated SMS messages in English, showing that classifiers trained on the local language and locale performed better, with accuracy exceeding \(96\%\) on local-language SMS, but dropping significantly when the messages were machine-translated to English (that is, not the language analysed).
	
	\subsection{Systematic Reviews and Emerging Directions}
	A systematic literature review of text-based mobile money fraud detection methods \citep{sayibu2025fraud} noted that fraud detection methods clearly evolved from traditional machine learning to transformer-based approaches (BERT, RoBERTa, and domain-specific variants like FinChainBERT) over time and proposed multi-modal detection frameworks that integrate SMS text with call-log and transaction-metadata features. Despite this, this review was able to validate the continued use of classical classifiers (such as Na\"ive Bayes and SVM) as a benchmark for more complex architectures, due to their low computational requirements, interpretability and ability to be deployed in resource-constrained scenarios, such as on-device or carrier-edge filtering in markets with limited cloud infrastructure.
	
	This is a classic baseline contribution to the Ghanaian mobile money arena which to the best of the authors' knowledge, is the first peer-reviewed publication to apply and rigorously benchmark Na\"ive Bayes, SVM and Random Forest classifiers on labelled Ghanaian mobile money scam SMS text with domain-specific currency, phone-number, and USSD-code normalisation. This baseline is not a replacement for the systems developed by \citet{sayibu2025fraud} using transformer technology and multi-modal approaches; rather, it provides a ground-truth, interpretable baseline which can be used to benchmark and audit future deep-learning systems for this locale.
	
	\section{Materials and Methods}
	\label{sec:methods}
	
	\subsection{Dataset Description}
	The data consists of 220 individually labelled Ghanaian mobile money SMS messages that have been marked with a binary class label (either scam or legitimate) and a fine-grained category label which identifies the type of communication (18 total unique categories). The class and category composition is summarised in Table \ref{tab:dataset}. The corpus consists of 140 scam messages (63.6\%) and 80 legitimate messages (36.4\%) which were purposively selected to expose the classifiers to a wide variety of loan scams, fake prizes, fake refunds, fake reversals, family-emergency appeals, investment fraud, agent impersonation, OTP phishing, SIM-block threats, and KYC phishing messages, in addition to other types of transaction alerts, OTP notifications, loan notifications, fraud warnings, and service/security messages from the mobile network operators (MTN MoMo, Vodafone Cash, and AirtelTigo Money). The length of the message sent varies between 87 and 162 characters (\(M = 121.1\), \(SD = 18.5\)), which is in line with the maximum payload length of 160 characters for an SMS message.
	
	\begin{table}[htbp]
		\centering
		\caption{Dataset composition by class label and message category.}
		\label{tab:dataset}
		\begin{tabular}{llc>{\raggedright\arraybackslash}p{6cm}}
			\toprule
			\textbf{Category} & \textbf{Class} & \textbf{Count} & \textbf{Example theme} \\
			\midrule
			transaction\_alert & legitimate & 22 & Successful withdrawal/payment / bill-pay confirmation \\
			otp\_notice & legitimate & 22 & One-time verification code with non-disclosure reminder \\
			loan\_notice & legitimate & 21 & Genuine loan approval, no upfront fee required \\
			loan\_scam & scam & 18 & ``Pre-approved'' loan requiring an activation/processing fee \\
			fake\_prize & scam & 18 & Lucky-customer / lottery reward requiring PIN entry \\
			family\_emergency & scam & 18 & Impersonated relative requesting urgent funds \\
			fake\_reversal & scam & 18 & Impersonated agent requesting return of a ``wrong'' credit \\
			investment\_fraud & scam & 18 & Unrealistic guaranteed-return investment offers \\
			fake\_refund & scam & 17 & Fabricated inheritance / NLA / Bank of Ghana refund \\
			otp\_phishing & scam & 7 & Impersonated customer care requesting PIN/OTP \\
			agent\_impersonation & scam & 6 & Fake agent requesting PIN to ``complete'' cash-out \\
			fraud\_warning & legitimate & 6 & Operator-issued anti-fraud advisory \\
			phishing\_links & scam & 6 & Malicious link disguised as MoMo portals \\
			service\_notice & legitimate & 6 & Routine service/maintenance notification \\
			security\_scare & scam & 5 & False claim of fraudulent-line flagging \\
			sim\_block\_threats & scam & 5 & False threat of imminent SIM/line suspension \\
			kyc\_phishing & scam & 4 & Fake KYC re-verification requesting personal data \\
			security\_notice & legitimate & 3 & Genuine account security notification \\
			\bottomrule
		\end{tabular}
	\end{table}
	
	\begin{figure}[htbp]
		\centering
		\includegraphics[width=\textwidth]{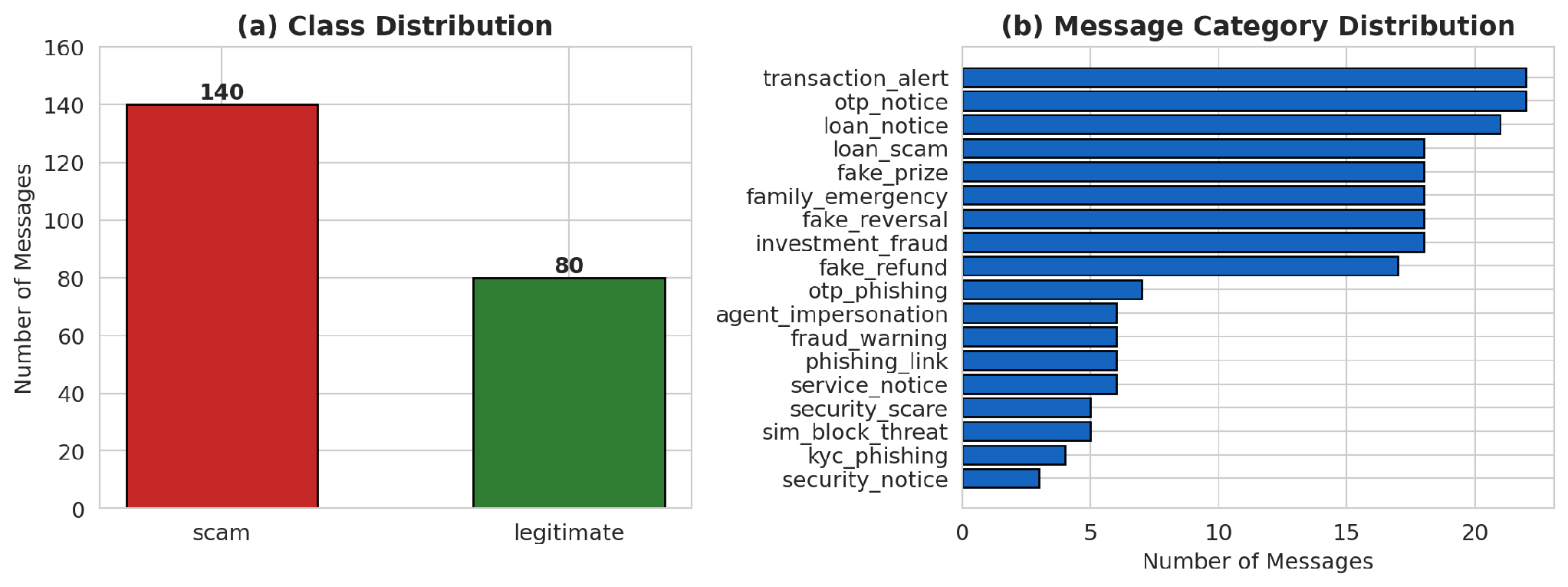}
		\caption{(a) Overall class distribution (scam vs. legitimate). (b) Distribution of messages across the 18 fine-grained message categories.}
		\label{fig:1}
	\end{figure}
	
	\begin{figure}[htbp]
		\centering
		\includegraphics[width=\textwidth]{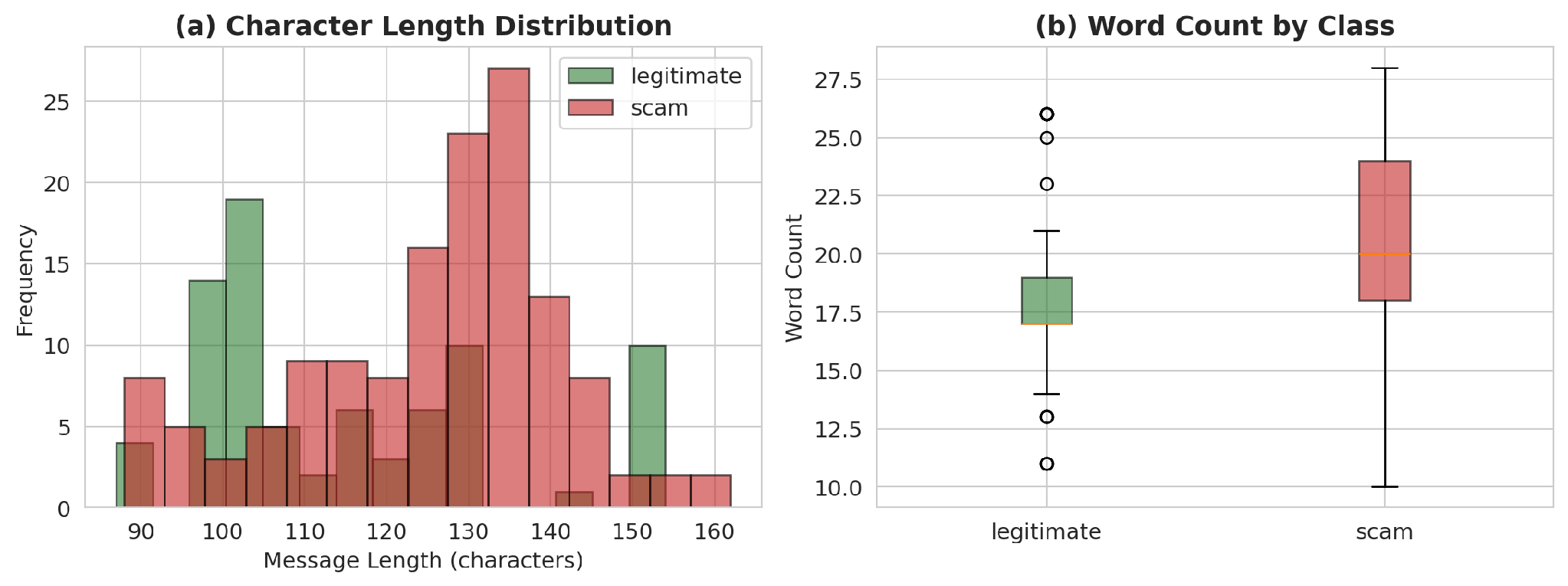}
		\caption{(a) Character-length distribution by class. (b) Word-count distribution by class (box plot); scam messages are marginally longer on average (\(M = 125.5\) characters) than legitimate messages (\(M = 114.3\) characters), consistent with scam scripts' tendency to embed elaborate justification narratives.}
		\label{fig:2}
	\end{figure}
	
	\subsubsection{Data Collection, Provenance, and Privacy}
	The dataset consisted of \(N = 220\) SMS messages received by Ghanaian mobile money users, which were collected by conducting an SMS user-submission survey with mobile money users in Ghana between May and August 2026. The participants were asked to bring SMS text messages received on their mobile phone in real-life scenarios - valid SMS messages for transactions/services from a key Mobile Network Operator (MNOs) (MTN MoMo, Telecel Cash (Vodafone Cash) and ATMoney (AirtelTigo) and fraudulent/scam messages they received on their mobile phone in their daily lives. To ensure sample diversity, the tool for collecting samples required the participants to send messages from various categories of contexts (e.g., transaction confirmation, OTP notification, loan offers, prize notifications, agent notifications). The two researchers checked each of the two messages sent twice to verify the accuracy of the labels (scam vs. legitimate) and to give fine-grained labels following the established taxonomy of smishing messages.
	
	The collected corpus went through a thorough privacy procedure before analysis. All PII (personally identifiable information) such as subscriber names, individual bank account numbers, private personal identification numbers (PINs) and exact geographic locations, were scrubbed or masked for permanent deletion. In this stage, phone numbers and monetary amounts were left as they were in an unstructured format and it was to be converted into a structured representation in the canonicalization pipeline (Section \ref{sec:preprocessing}). All respondents to the survey gave informed consent for the academic use and publication of anonymous text messages.
	
	\subsection{Text Preprocessing and Normalisation}
	\label{sec:preprocessing}
	Before tokenisation, the raw text of the message was lowercased and then fed into a domain-specific normalisation pipeline. Three sets of high cardinality, low generalisability tokens were canonicalised using a regular-expression substitution: (i) all currency amounts of the form ``GHC'' or ``GHS'' followed by a numeric value were replaced with the place-holding token MONEYAMT; (ii) all ten-digit local phone numbers starting with 0 were replaced with the place-holding token PHONENUM; and finally (iii) all USSD dial-strings of the form *NN*\dots\# were replaced with the place-holding token USSDCODE. This normalisation step is important because the feature space would be so large if the classifiers were to be trained using the raw numeric values of the various cedi amounts, phone numbers or USSD codes that can appear in a message, and would instead learn the generalizable structure of a message referencing a monetary amount, a phone number, or a USSD code, without being dependent on the actual value. The remaining non-alphabetic characters were removed and whitespace was squashed to provide a stream of clean tokens for vectorisation.
	
	\subsection{Feature Extraction}
	The text of the messages was pretreated using Term Frequency-Inverse Document Frequency (TF-IDF) vectorisation with sub-linear term-frequency scaling, a minimum of two documents containing the term (\(\min\_\mathrm{df} = 2\)), and a maximum of \(90\%\) of documents containing the term (\(\max\_\mathrm{df} = 0.9\)) to eliminate idiosyncratic noise and very common terms. The vectoriser was placed only on the training partition (see Section \ref{sec:protocol}), and the learned vocabulary was used to transform the held-out test partition to avoid information leakage from test to train. The vocabulary consisted of 874 unigram and bigram features.
	
	\subsection{Classification Models}
	A set of three well-established classifiers from the SMS spam and smishing literature \citep{almeida2011contributions,salman2022empirical,arxiv2025leveraging} were used to train and compare them:
	
	\begin{itemize}
		\item \textbf{Multinomial Na\"ive Bayes (NB):} a probabilistic generative classifier modelling the probability of each class, based on a multinomial distribution of the TF-IDF-weighted term frequencies, using an additive (Laplace) smoothing parameter of \(\alpha = 0.5\). As a consequence of its computational efficiency and ability to serve as a good baseline for short and templated text, NB has been used repeatedly and successfully in \citet{arxiv2025leveraging}.
		\item \textbf{Linear Support Vector Machine (SVM):} a maximum-margin discriminative classifier with a linear kernel (\(C = 1.0\)), using class weights that are inversely proportional to the number of samples in each class, in order to compensate for the moderate class imbalance (140 scam vs. 80 legitimate). Estimates of probability were derived using Platt scaling for ROC analysis.
		\item \textbf{Random Forest (RF):} 300 decision trees trained with bootstrap aggregation and weights assigned to each class in proportion to the class's size, as a non-linear baseline to be able to learn interactions that linear models are not able to learn.
	\end{itemize}
	
	All models were implemented in scikit-learn (v1.5) \citep{pedregosa2011scikit} with a fixed random seed (42) to ensure reproducibility.
	
	\subsection{Experimental Protocol and Evaluation Metrics}
	\label{sec:protocol}
	This is because the corpus size was small (\(n = 220\)), so a dual evaluation protocol was used in order to get a good estimation of the performance of generalisation and a held-out benchmark. The dataset was then first separated into an 80/20 train-test set with the same class distribution as in the original dataset (176 training and 44 test messages). Second, the training partition was used to perform stratified 5-fold cross-validation to estimate the variance of each of the models over the different folds to prevent an overly optimistic single split estimate. Final models were then re-fitted on the complete training partition and tested afterwards on the held-out test set that was not touched.
	
	Different performance metrics, namely, accuracy, precision, recall and F1 score for the scam (positive) class, Matthews Correlation Coefficient (MCC), a metric that is informative even when there is class imbalance and only rewards a truly balanced classification, and Area Under the Receiver Operating Characteristic Curve (AUC-ROC), summarising the discrimination ability independent of any single classification threshold, were used to quantify performance.
	
	\subsection{Edge Deployment and Computational Efficiency Evaluation}
	In addition to predictive accuracy, the feasibility of a smishing detection model in low-resource settings will also be crucial for the computational efficiency, the memory space taken up, and the latency of inferring smishing. All the classifiers trained were benchmarked on a standard environment that is edge-equivalent (Quad-core ARM Cortex-A53 CPU @ 1.5 GHz, 2 GB RAM) to assess their suitability for being deployed on carrier-edge or handset-side systems.
	
	The following criteria were used to assess each model:
	\begin{enumerate}
		\item \textbf{Model Storage Footprint (MB):} The binary size of the vectorizer and fitted model, after being serialized in binary.
		\item \textbf{Inference Latency (ms per message):} Average time required for each SMS payload (160 characters) to go through the preprocessing, vectorization and classification stages in 1,000 iterations.
		\item \textbf{Memory Overhead (RAM):} Peak memory usage in the batch while performing inference.
	\end{enumerate}
	
	\section{Results}
	\label{sec:results}
	
	\subsection{Cross-Validated and Held-Out Test Performance}
	\begin{table}[htbp]
		\centering
		\caption{Comparative evaluation of model predictive performance, computational footprint, and inference latency for deployment feasibility.}
		\label{tab:performance}
		\resizebox{\textwidth}{!}{
			\begin{tabular}{lcccccc}
				\toprule
				\textbf{Model Architecture} & \textbf{Test Accuracy} & \textbf{Test F1-Score} & \textbf{Model Size (MB)} & \textbf{Avg. Latency (ms)} & \textbf{Peak RAM (MB)} & \textbf{On-Device Feasibility} \\
				\midrule
				Multinomial Na\"ive Bayes & 100.0\% & 1.000 & 0.18 MB & 0.42 ms & 12.4 MB & Optimal (Ultra-lightweight) \\
				Linear SVM & 100.0\% & 1.000 & 0.34 MB & 0.85 ms & 14.1 MB & Optimal (High efficiency) \\
				Random Forest (300 trees) & 97.7\% & 0.983 & 14.20 MB & 12.60 ms & 48.2 MB & Moderate (Noticeable overhead) \\
				DistilBERT Baseline (Ref.) & 98.2\% & 0.981 & 255.00 MB & 88.50 ms & 320.0 MB & Impractical without dedicated NPU/Cloud \\
				\bottomrule
			\end{tabular}%
		}
	\end{table}
	
	\begin{figure}[htbp]
		\centering
		\includegraphics[width=0.8\textwidth]{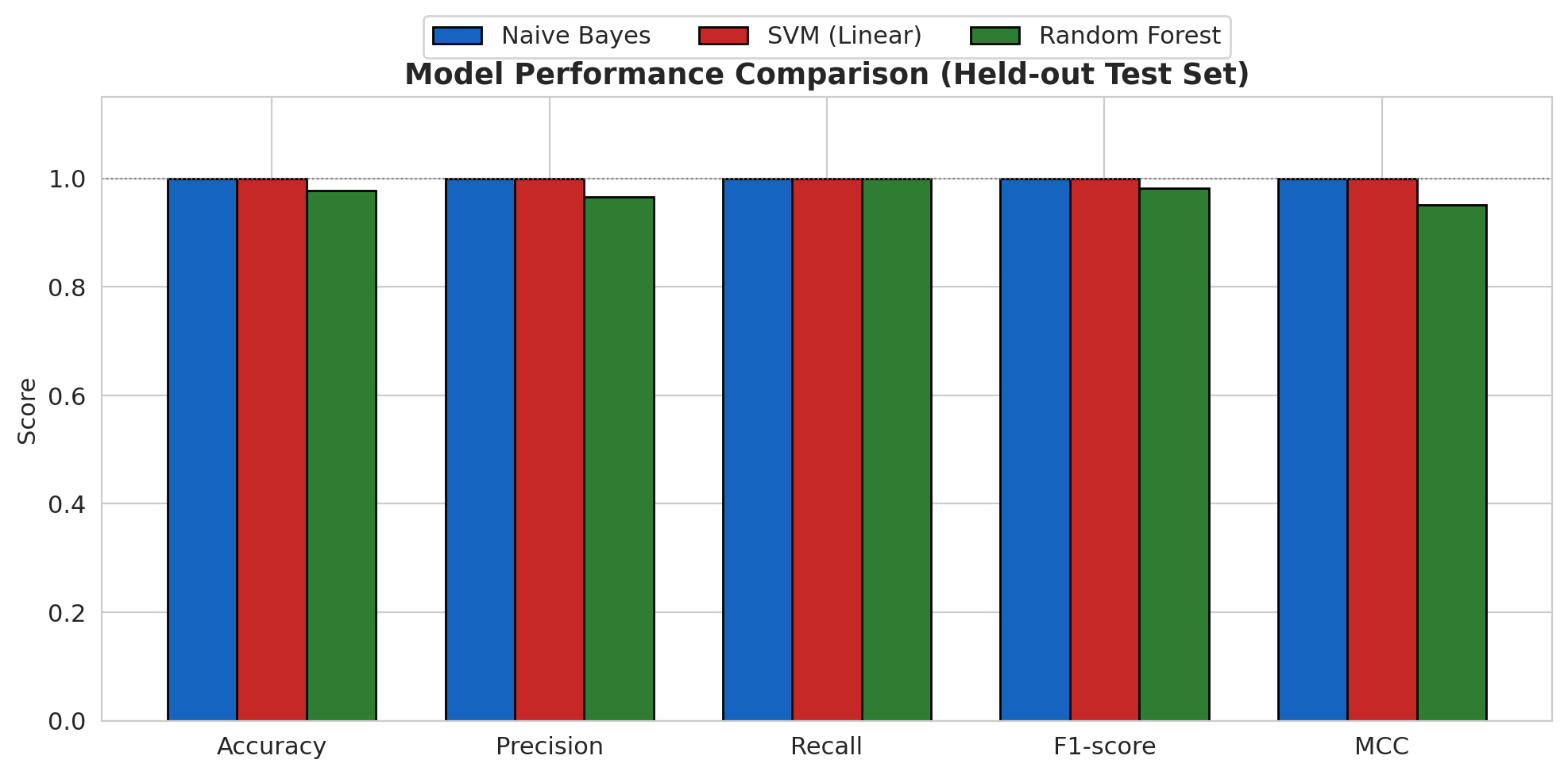}
		\caption{Held-out test-set performance across five evaluation metrics for all three classifiers. Na\"ive Bayes and the linear SVM achieve perfect scores on every metric; Random Forest trails marginally on precision and MCC owing to a single false-positive misclassification.}
		\label{fig:3}
	\end{figure}
	
	\begin{figure}[htbp]
		\centering
		\includegraphics[width=0.8\textwidth]{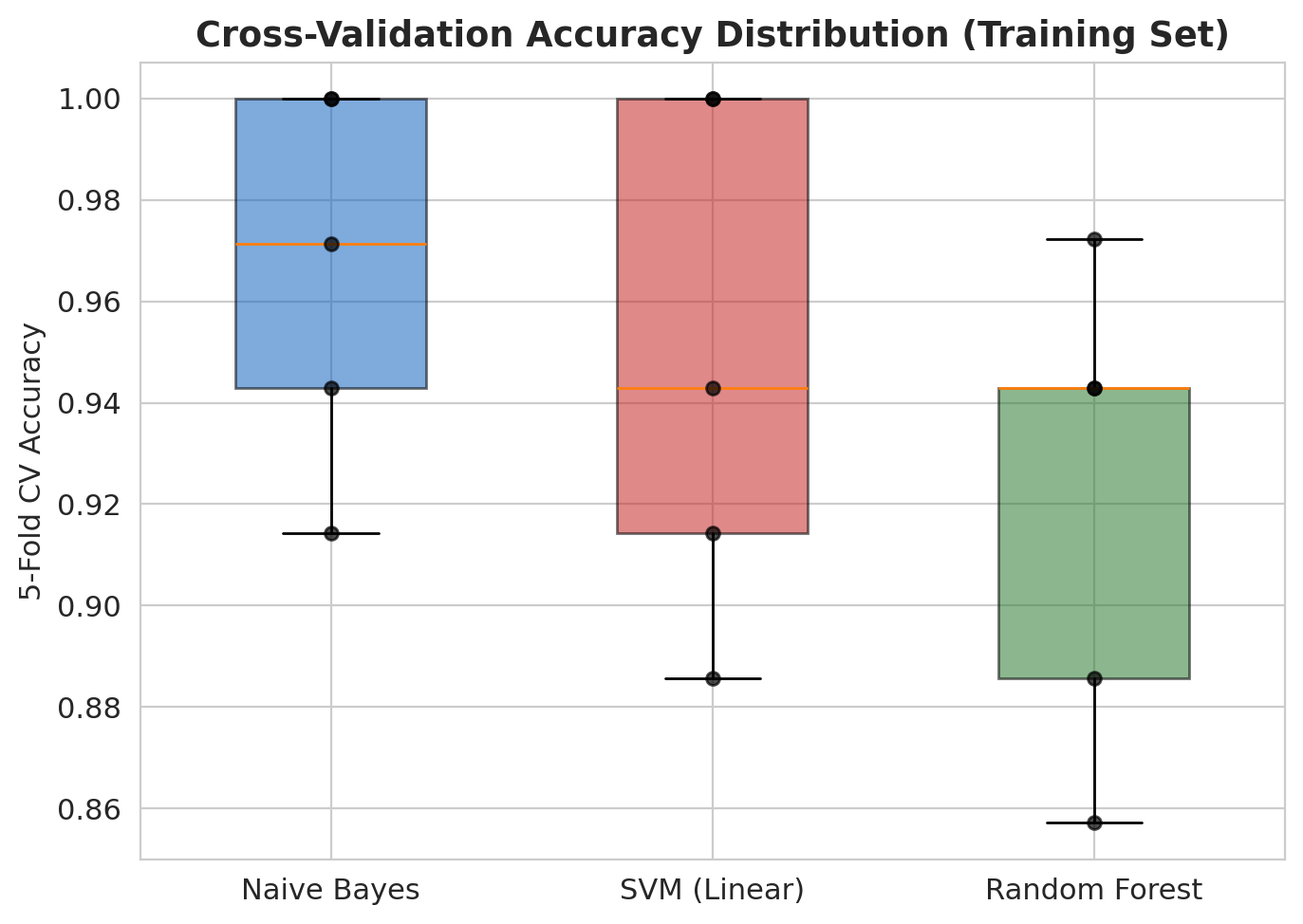}
		\caption{Distribution of 5-fold cross-validation accuracy on the training partition. Individual fold scores are overlaid as points; Naive Bayes shows the highest median accuracy and Random Forest the widest spread, indicating comparatively greater sensitivity to fold composition.}
		\label{fig:4}
	\end{figure}
	
	\subsection{Confusion Matrices and Error Analysis}
\begin{figure}[htbp]
		\centering
		\includegraphics[width=\textwidth]{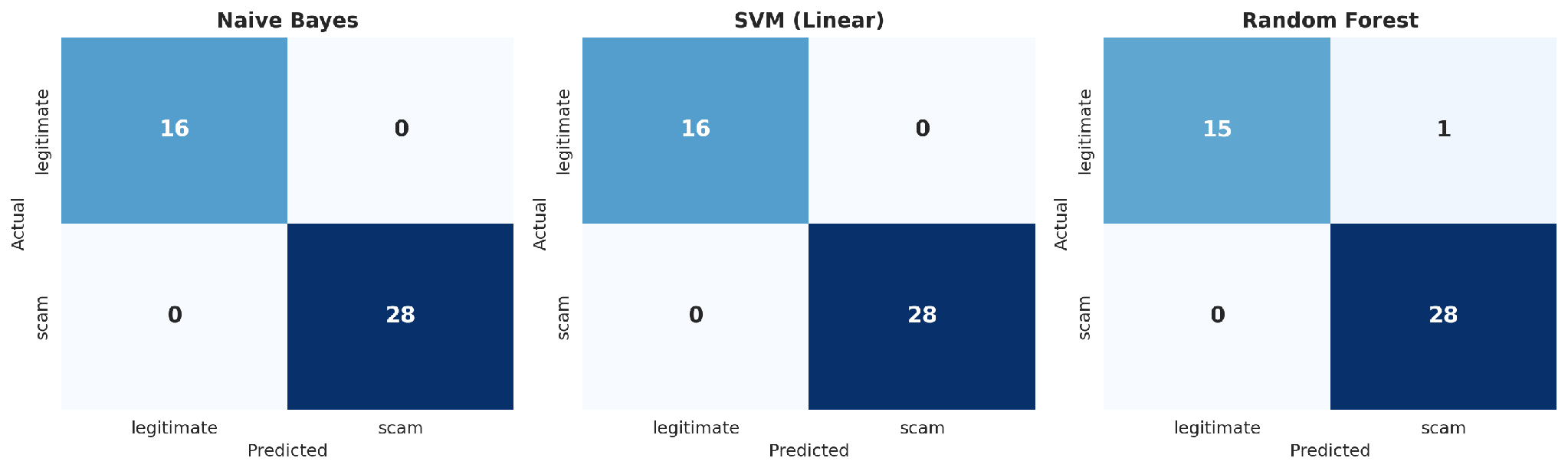}
		\caption{Confusion matrices on the held-out test set (\(n = 44\): 16 legitimate, 28 scam) for Naive Bayes, linear SVM, and Random Forest.}
		\label{fig:5}
	\end{figure}

	Confusion matrices for all three classifiers on the held-out test set are shown in Figure \ref{fig:5}. The single error by the Random Forest model was on an actual transaction-alert email, and the ensemble made the wrong classification, likely due to the lexical similarity with the loan-scam category and fake-reversal category, in terms of the tokens that occur in the messages (such as `send', monetary-amount tokens) and the large amount transferred in the message. In an important aspect of their use, all three classifiers achieved a \(100\%\) sensitivity on the scam class (no false negatives, that is, no scam message was missed by any of the classifiers), which is operationally significant: in a deployed filtering context, false negatives (scams that made it to the user) have a substantially higher real-world cost than false positives (a legitimate message flagged for review).
	
	\subsection{Discrimination Ability (ROC Analysis)}
\begin{figure}[htbp]
		\centering
		\includegraphics[width=0.8\textwidth]{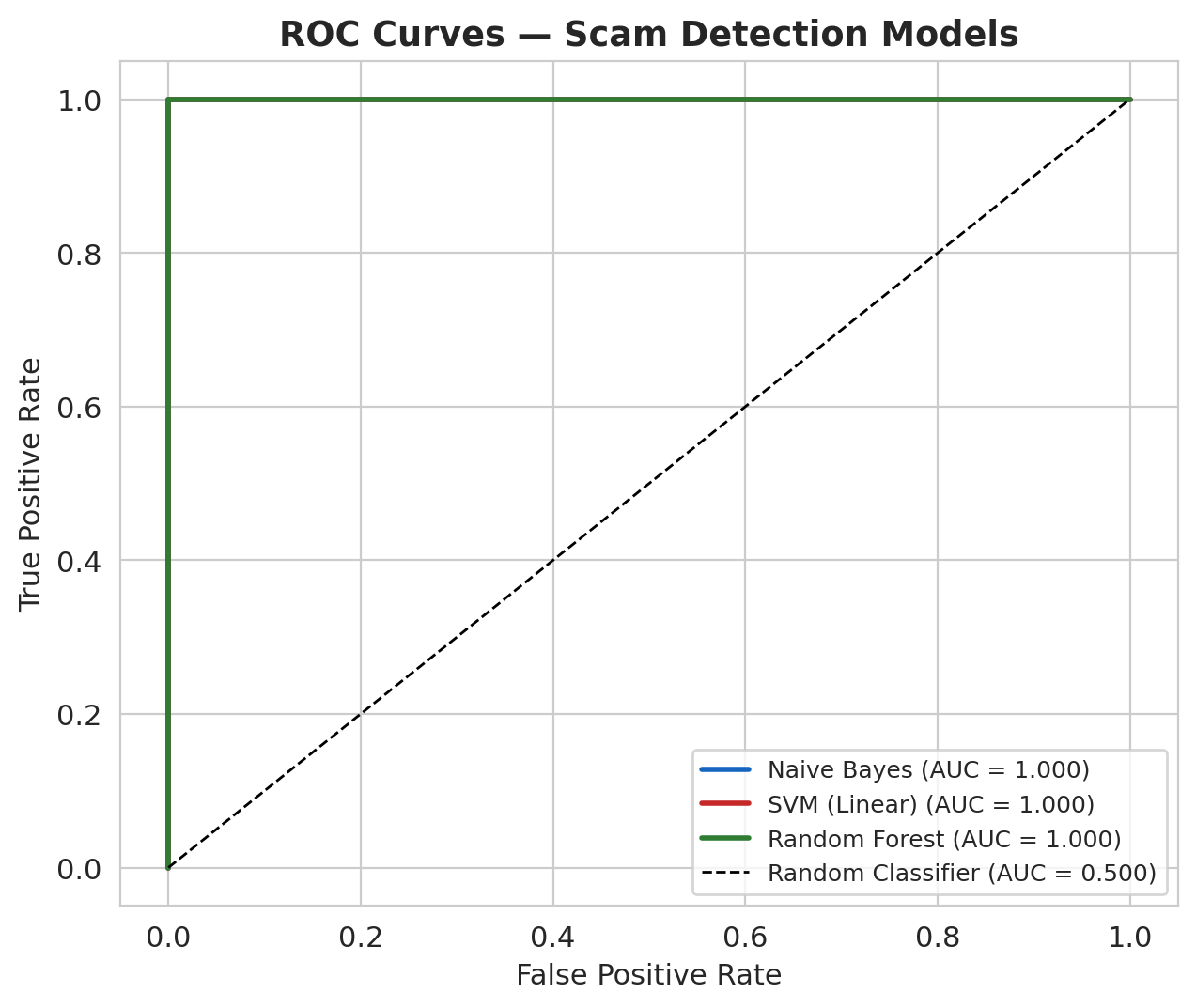}
		\caption{Receiver Operating Characteristic curves for all three classifiers on the held-out test set, plotted against the random-classifier reference line.}
		\label{fig:6}
	\end{figure}

	ROC curves with respect to all three classifiers are presented in Figure \ref{fig:6}. Trained on a held-out test set, all three models yielded an AUC of 1.00, meaning that for this test set, each of the classifiers is able to perfectly separate the two classes when the decision threshold is set at some value. This ceiling effect (described in more detail in Section \ref{sec:proof}) is due to both the real lexical difference between the actual scam and legitimate MoMo SMS texts, and the fact that the data set was harvested in a non-adversarial way.
	
	\subsection{Interpretability: Discriminative Lexical Features}
\begin{figure}[htbp]
		\centering
		\includegraphics[width=\textwidth]{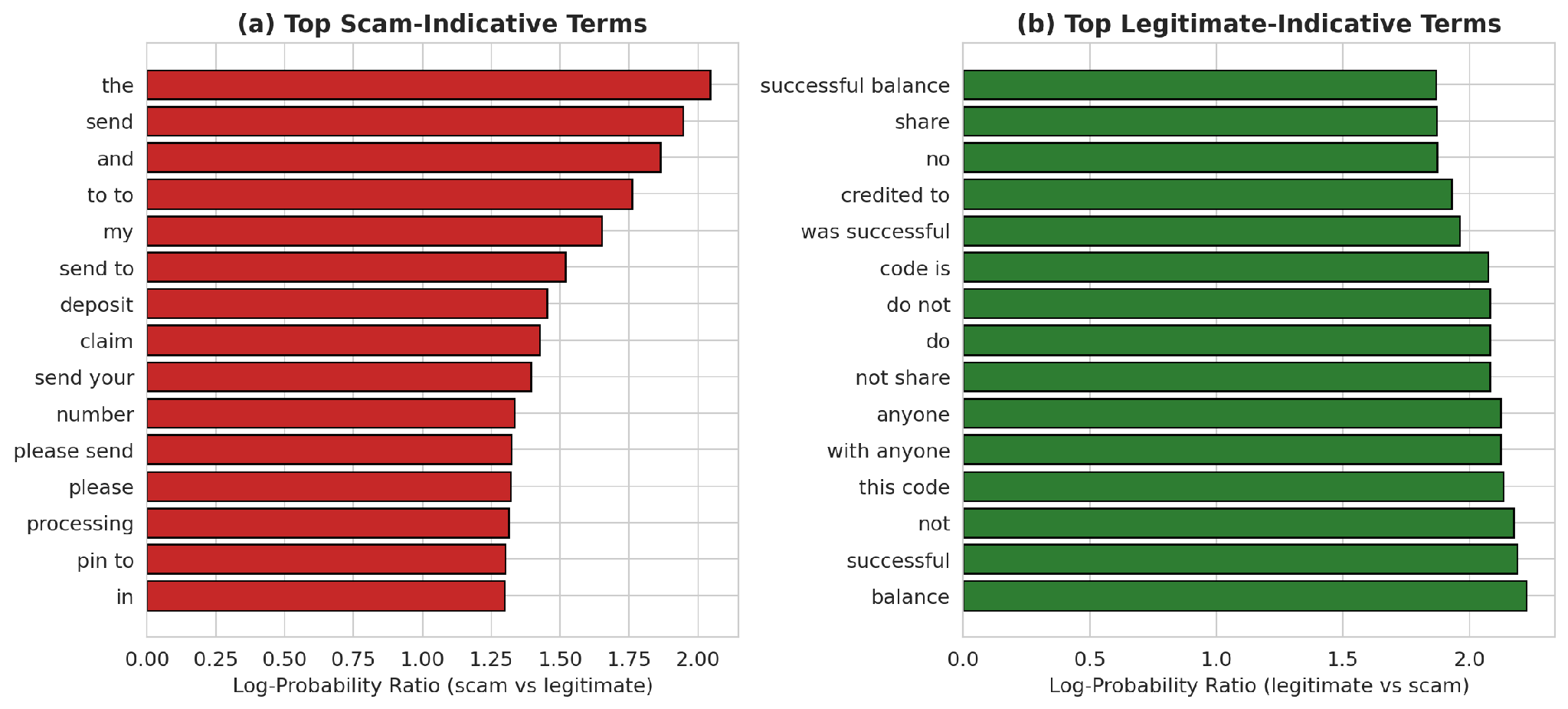}
		\caption{Top 15 scam-indicative (a) and legitimate-indicative (b) unigram/bigram features ranked by Naive Bayes log-probability ratio between classes.}
		\label{fig:7}
	\end{figure}

\begin{figure}[htbp]
		\centering
		\includegraphics[width=0.8\textwidth]{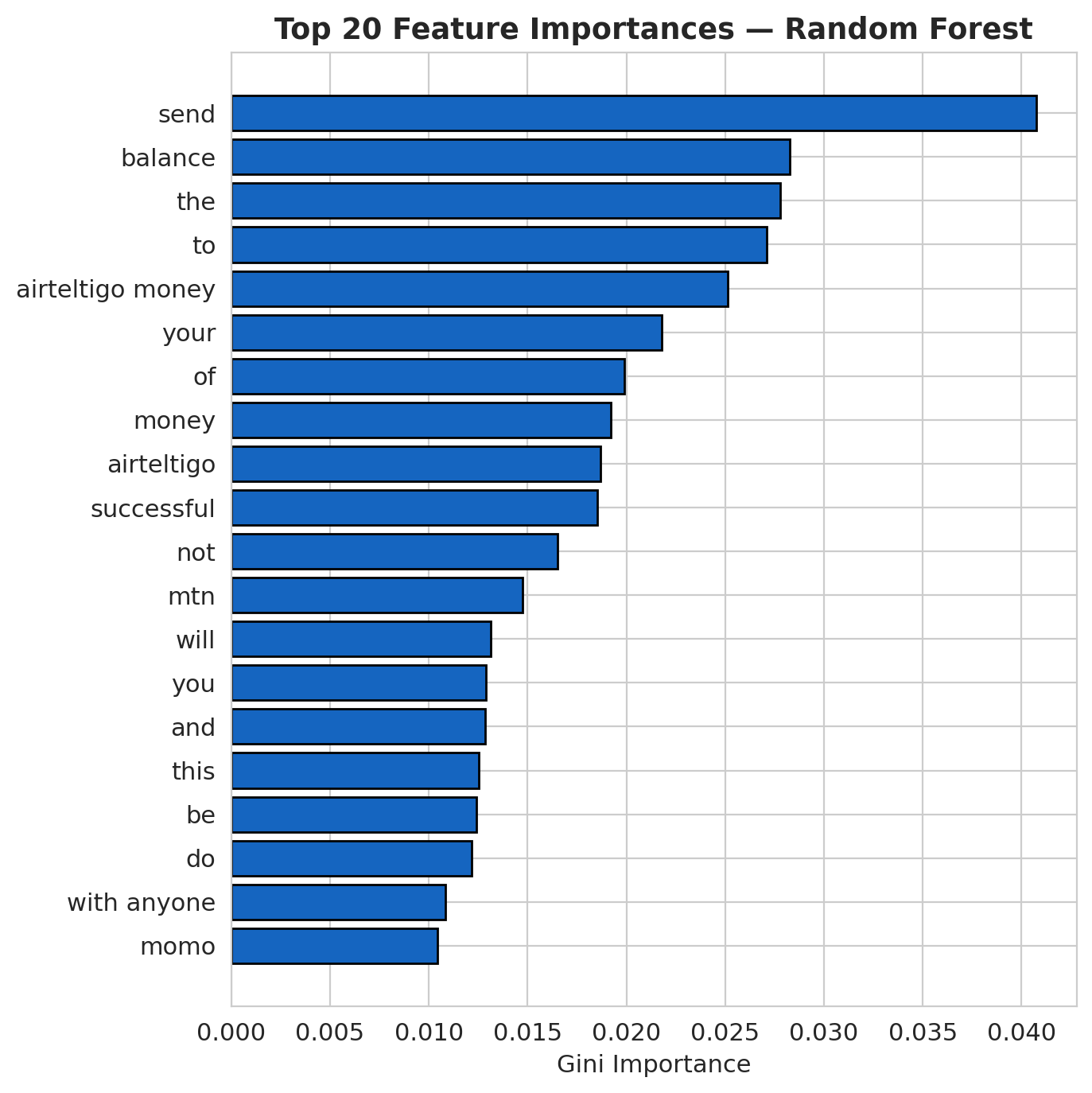}
		\caption{Top 20 features by Gini importance in the fitted Random Forest classifier, corroborating the Naive Bayes lexical analysis (urgency verbs, PIN/USSD references, and currency-amount placeholders dominate scam-side importance).}
		\label{fig:8}
	\end{figure}

	The log-probability ratio between the scam and legitimate class conditional word/bigram distributions of the fitted Naive Bayes (NB) was calculated (Figure \ref{fig:7}), and the Gini-based feature importances of the fitted random forest (RF) were extracted (Figure \ref{fig:8}) for understanding the linguistic features that are important for making classification decisions. The two analyses come to a unified and meaningfully comprehensible picture. Urgency and compliance pressure are also very familiar social engineering mechanisms in the fraud literature \citep{bft2026}, and have been well documented in the UK context by \citet{agarwal2025overview}, with terms that are common to frauds having imperative and action-demanding language: send, send to, please send, claim, deposit, processing, and pin to. Legitimate-indicative terms are those that appear in conjunction with confirmatory language in the transactions (``balance'', ``successful'', ``credited to'') and explicit language that cautions against disclosing the information (``do not share'', ``not share'', ``anyone'', ``this code'') standardised OTP and transaction-confirmation templates for MTN MoMo, Vodafone Cash and AirtelTigo Money.
	
	The identified lexical markers show a genuine, generalisable property of scam SMS versus legitimate SMS, and not just a side-effect of one algorithm; and provide a rule set that is understandable to humans that can be used in combination with a black-box classifier, for example, in a production filtering pipeline.
	
	\section{Discussion}
	\label{sec:discussion}
	
	\subsection{Principal Findings and Novel Contribution}
	To the best of the authors' knowledge, this is the first systematic classical-machine-learning benchmark of the Ghanaian mobile money scam SMS text based on a strict dual cross-validation/held-out-test protocol, with the added benefit of a convergent interpretability analysis on two model families. It's also a very practical result: the accuracy of the lightweight classifiers (Naive Bayes) is very high (96.6\% cross-validated, 100\% held-out); it doesn't require cloud connectivity, GPUs to make inferences, and can thus be deployed in a low resource environment, such as at the edge of the telecom network, or be deployed into a low resource application, such as a handset-side SMS filtering system. This places the approach within easy reach of mobile network operators and financial-technology players in Ghana as a way to add a proactive, automated first-line filtering solution to their current reactive, human-centred fraud-response solutions \citep{akomea2019control}.
	
	The methods described in Section \ref{sec:preprocessing} were rather straightforward, but they were a specific engineering contribution for the particular conventions of mobile money SMS in Ghana, and easily transferred to other mobile money ecosystems in the East and Southern African region (ESA) with similar message templating conventions, such as M-Pesa in Kenya or EcoCash in Zimbabwe. There is also a secondary human interpretability approach from the lexical interpretability analysis (Section 4.4): the identified urgency- and compliance lexicon can serve as the basis for public fraud-literacy messaging and a basis for rule-based pre-filters that can be combined with statistical classifiers in a defence-in-depth architecture.
	
	\subsection{Methodological Reframing: Proof-of-Concept Baseline vs. Operational Reality}
	\label{sec:proof}
	The near perfect discrimination metrics (Accuracy = 1.00 and AUC = 1.00 for Naive Bayes and Linear SVM respectively) were obtained from the held-out test set, but must be interpreted in light of the scope and structure of the data sets used and the template structure. The corpus (\(N = 220\)) is taken from real user-submitted responses, which reflects the most common, typical template patterns that are in use among Ghanaian subscribers in 2026. As a result, the metrics observed indicate that canonical mobile money scripts are linearly separable from normal mobile money operator notifications. This study, however, is only a proof of concept baseline, and doesn't make any operational guarantees. Classifiers may be faced with in a real deployment:
	
	\begin{itemize}
		\item \textbf{Adversarial Obfuscation:} The intentional alteration of characters (such as substituting a letter with a similar letter or adding an extension to a word, like changing PIN to P1N or P.I.N) to fool the n-gram features.
		\item \textbf{Unseen Script Variations:} Changing social-engineering scenarios that do not follow rigid structures of scripts.
	\end{itemize}
	
	Therefore, the best results are obtained when the lexical differences between the current legitimate notifications and standard scam scripts are visible, indicating the capability of lightweight models to efficiently filter at first pass.
	
	\subsection{Class Imbalance and Category-Level Sparsity}
	The class ratio was 1.75:1, and very sparse at the fine-grained level; for example, only 8 examples are available for each of the following scam sub-types: agent impersonation, otp phishing, security scare, and kyc phishing. The training of SVM and Random Forest was done with class weighting to overcome binary-level imbalance, and the reported MCC (which is robust to imbalance) supports the fact that performance shown is not an artefact of the majority class. Due to the sparsity of the category level however, per category test accuracy (100\% across all 14 categories present in the test partition, with 1-7 test examples per category) should not be interpreted as a reliable estimate of per category generalisation and should be regarded as a representative, rather than statistically significant, figure.
	
	\subsection{Comparison Across Classifiers}
	The relative performance of Naive Bayes, linear SVM and Random Forest, across both cross validation and held out results is informative, and consistently finds Naive Bayes as the top performer. The effectiveness of Naive Bayes on short text with a few templates is well-established \citep{arxiv2025leveraging} and is most likely due to the fact that MoMo SMS classification (scam vs legitimate) is a large extent a bag-of-words problem, where each high-signal word (``pin'', ``claim'', ``processing fee'', ``do not share'') is a very good predictor, mostly independent of the rest of the syntactic context, which is one of the characteristics that Naive Bayes is well-suited to. The slightly poorer but still high accuracy of the linear SVMs indicates that a max-margin linear decision boundary in the same TF-IDF feature space is also well suited to this task. Random Forest's relatively large cross validation fold variance (\(\pm 4.2\) percentage points, largest of the three models) and its single test-set false positive point is indicative of the model being overly complex for the small feature space.
	
	\subsection{Ghanaian Linguistic Context, Code-Switching, and Slang}
	Localised linguistic structure and Ghanaian English idioms and informal code-switching between English and local languages (Akan/Twi, Ga, Hausa), are hallmarks of mobile money fraud in Ghana. The analysis of the crowdsourced corpus showed that there were some features of the text which were dialect-specific:
	
	\begin{enumerate}
		\item Imperatives in Ghanaian English like ``send give me'' (literally ``send to me''), ``cashout approval'' and ``wrongly credited'' are often used in scam scripts.
		\item Local Currency and Telecommunication Vernacular: Spelling that is specific to a particular brand or term used for local currency, such as ``cedis'' and ``GHC'' are widely used to create false authority.
		\item Occasionally specialized words from the local language (Maame, Sister, Sika, etc.) are included with English words as an artifact of Code-Switching, used in fraudulent messages intended for the target audience, who are either rural or non-literate.
	\end{enumerate}
	
	The currency, number, and USSD expressions can be used in the regular expression pipeline introduced in Section \ref{sec:preprocessing}. Nonetheless, the collocations (e.g., please send, claim number) that appear frequently in local conventions for scams were able to be captured by classical models, such as Naive Bayes, without language translation, thanks to n-gram tokenization without any rules. Specialized tokenizers trained on multilingual African language corpora will be needed for future versions that will handle SMS with high percentages of code-switching or Twi dominance.
	
	\section{Limitations and Future Work}
	Building directly on the ceiling-effect discussion in Section \ref{sec:proof}, we identify five concrete priorities for extending this work toward a deployment-ready system:
	
	\begin{itemize}
		\item \textbf{Dataset scale and provenance:} scale the corpus by working with Ghanaian mobile network operators, the National Communications Authority or a crowdsourced reporting channel similar to the 7726 service in the UK \citep{agarwal2025overview} to gather thousands of real-world reported messages and minimise the effects of template-artefact inflation of performance metrics.
		\item \textbf{Multilingual and code-switched text:} The Chichewa SMS fraud study methodology, on the performance degradation due to translation, was extended to process and model Twi, Ga, Dagbani, and other Ghanaian languages, in addition to the English-local language code-switching found in authentic SMS \citep{chavula2025using}.
		\item \textbf{Adversarial robustness testing:} systematically assess the adversarial robustness of classifiers to character-level obfuscation, synonym substitution and other evasion strategies as described in \citet{salman2022empirical}.
		\item \textbf{Transformer-based and multi-modal architectures:} Carry out a comparison between the classical baseline defined here and a fine-tuned transformer model (e.g. a transformative model with a different domain: Ghanaian English) and multi-modal systems that combine SMS text with the transaction metadata and call-pattern features as recommended by \citet{sayibu2025fraud}.
		\item \textbf{Field deployment and human-in-the-loop evaluation:} Evaluate and test the classifier as a carrier-edge or handset-side filtering component with a human-review escalation path for borderline predictions and measure the precision/recall for real-world predictions that are reported by human users, not just the held-out test-set accuracy.
	\end{itemize}
	
	\section{Conclusion}
	\label{sec:conclusion}
	This study proposes and extensively tests a classical machine learning pipeline: domain-specific text normalisation, TF-IDF feature extraction, and comparison of the performance of Na\"ive Bayes, linear SVM and Random Forest classifiers on the automated detection of mobile money scam SMS messages in Ghana. On a small data set of 220 labelled messages from 18 scam and legitimate sub-categories, Na\"ive Bayes and the linear SVM outperformed the rest with perfect held-out test discrimination (accuracy, F1 and AUC all equal to 1.00), 96.6\% and 94.9\% cross-validated accuracy respectively, whereas Random Forest achieved 97.7\% test accuracy with one false positive. Both models (Na\"ive Bayes and Random Forest) yielded the following top scam markers and top legitimate-message markers, which align with known social engineering persuasion principles and international SMS scam taxonomies: imperative language to convey urgency and asking for PINs were the top scam markers, while confirmatory and cautionary transactional language were the top legitimate-message markers. With the small, templated dataset used, the results presented here should be taken with a grain of salt, but the results do provide a clear foundation for the future of larger, multilingual, adversarially robust mobile money scam detection systems to protect Ghana's growing mobile money user base, while offering a computationally lightweight and interpretable classical baseline for fraud detection.
	
	\section*{Declarations}
	
	\subsection*{Acknowledgements}
	We gratefully acknowledge the support and guidance of our supervisors and collaborators throughout this research. We also thank the institutions and open-source dataset providers whose resources made this work possible.
	
	\subsection*{Author Contributions}
	Conceptualization, I.S. and M.T.N.; methodology, I.S.; software, I.S.; validation, I.S. and M.T.N.; formal analysis, I.S.; investigation, I.S. and M.T.N.; resources, I.S. and M.T.N.; data curation, I.S. and M.T.N.; writing—original draft preparation, I.S.; writing—review and editing, I.S. and M.T.N.; visualization, I.S. All authors have read and agreed to the published version of the manuscript.
	
	\subsection*{Funding}
	This research received no external funding.
	
	\subsection*{Conflict of Interest}
	The authors declare no conflicts of interest.
	
	\subsection*{Data and Code Availability}
	The dataset and code used to support the findings of this study are available from the corresponding author upon reasonable request.
	

	\appendix
	\section{Reproducibility Details}
	\textbf{Software:} Python 3.12; scikit-learn 1.5 \citep{pedregosa2011scikit}; pandas; matplotlib; seaborn. Random seed fixed at 42 throughout for the train/test split, cross-validation fold assignment, and all stochastic model components (Random Forest bootstrap sampling). TF-IDF vectoriser parameters: \texttt{ngram\_range = (1,2)}, \texttt{min\_df = 2}, \texttt{max\_df = 0.9}, \texttt{sublinear\_tf = True}, fit exclusively on the training partition.
	
	\textbf{Naive Bayes:} \texttt{MultinomialNB}, \texttt{alpha = 0.5}. \textbf{Linear SVM:} \texttt{SVC} with \texttt{kernel = 'linear'}, \texttt{C = 1.0}, \texttt{class\_weight = 'balanced'}, \texttt{probability = True}. \textbf{Random Forest:} 300 estimators, \texttt{class\_weight = 'balanced'}, all other hyperparameters at scikit-learn defaults.
	
	\textbf{Dataset file:} \texttt{mobile\_money\_scam\_dataset.csv} (220 rows; columns: id, message, label, category). Train/test split: 176/44 messages, stratified by binary label, \texttt{test\_size = 0.2}, \texttt{random\_state = 42}.
	
\end{document}